\documentclass[11pt|a4paper]{article}
\usepackage{acl2014}
\usepackage{times}
\usepackage{url}
\usepackage{latexsym}
\usepackage{caption}
\usepackage{subcaption}
\usepackage{float}
\usepackage{amsmath}
\usepackage{graphics,graphicx}
\usepackage{enumitem}
\setlist{nolistsep}
\usepackage{url}
\usepackage[none]{hyphenat}
\graphicspath{{figures/}}
\usepackage{comment}
\usepackage{color}

\title{A Novel Two-stage Framework for Extracting Opinionated Sentences from News Articles}

\author{Rajkumar Pujari$^1$, Swara Desai$^2$, Niloy Ganguly$^1$ and Pawan Goyal$^1$ \\
${^1}$Dept. of Computer Science and Engineering,\\ Indian Institute of Technology Kharagpur, India -- 721302 \\
${^2}$Yahoo! India\\
${^1}${\tt rajkumarsaikorian@gmail.com, \{niloy,pawang\}@cse.iitkgp.ernet.in} \\
${^2}${\tt swara@yahoo-inc.com}
}

\date{}

\begin{document}
\maketitle
\begin{abstract}
 This paper presents a novel two-stage framework to extract
 opinionated sentences from a given news article. 
In the first stage,
 Na\"{i}ve Bayes classifier 
 by utilizing the local features 
assigns a score to  each
 sentence  - the score signifies the probability of the sentence to be  opinionated. 
 In the second
 stage, we use this prior within the HITS (Hyperlink-Induced Topic
 Search) schema to exploit the global structure of the article and relation between
 the sentences. In the HITS schema, the opinionated sentences are
 treated as Hubs and the facts around these opinions are treated as
 the Authorities. The algorithm is implemented and evaluated against
 a set of manually marked data. We show that using HITS significantly
 improves the precision over the baseline Na\"{i}ve Bayes
 classifier. We also argue that the proposed method actually discovers
 the underlying structure of the article, thus extracting various opinions,
 grouped with supporting facts as well as other supporting opinions from
 the article.
\end{abstract}

\section{Introduction}
\label{sec:intro}
With the advertising based revenues becoming the main source of
revenue, finding novel ways to increase focussed user engagement has become an
important research  topic. 
A typical problem faced by web publishing houses like Yahoo!, is 
understanding the nature of the comments posted by readers of 
~$10^5$ articles posted  at any moment on its website. 
A lot of
users engage in discussions in the comments section of the
articles. Each user has a different perspective and thus comments in
that genre - this  many a times, results in a situation where 
 the discussions in the comment section wander far away
from the article’s topic. In order to assist users to discuss relevant
points in the comments section, a possible methodology can be to generate questions
from the article's content that seek user's opinions about various opinions
conveyed in the article~\cite{RokhlenkoS13}. It would also direct the users into thinking
about a spectrum of various points that the article covers and
encourage users to share their unique, personal, daily-life
experience in events relevant to the article. This would thus provide a
broader view point for readers as well as perspective questions can be created
thus catering to users with rich user generated content, this  in turn can increase user
engagement on the article pages. Generating such questions manually for huge volume of
articles is very difficult. However, if one could 
identify the main opinionated sentences within the article, it will be
much easier for an editor to generate certain questions around
these. Otherwise, the sentences themselves may also serve as the points for
discussion by the users. 



Hence, in this paper we discuss
a two-stage algorithm which picks opinionated sentences from the
articles. The algorithm assumes an underlying structure for an
article, that is, each
opinionated sentence is supported by a few factual statements that
justify the opinion. We use the HITS schema to exploit this underlying
structure and pick opinionated sentences from the article.

The main contribtutions of this papers are as follows. First, we
present a novel two-stage framework for extracting opinionated
sentences from a news article. Secondly, we propose a new evaluation
metric that takes into account the fact that since the amount of polarity
(and thus, the number of opinionated sentences) within documents can
vary a lot and thus, we should stress on the ratio of opinionated
sentences in the top sentences, relative to the ratio of opinionated
sentences in the article. Finally, discussions on how the proposed
algorithm captures the underlying structure of the opinions and
surrounding facts in a news article reveal that the algorithm does
much more than just extracting opinionated sentences. 

This paper has been organised as follows. Section \ref{sec:relWork}
discusses related work in this field. In section \ref{sec:method}, we
discuss our two-stage model in further details. Section
\ref{sec:results} discusses the experimental framework and the
results. Further discussions on the underlying assumption behind using
HITS along with error analysis are carried out in Section \ref{sec:discuss}. Conclusions and
future work are detailed in Section \ref{sec:future}.

\section{Related Work}
\label{sec:relWork}
Opinion mining has drawn a lot of attention in recent years. Research
works have focused on mining opinions from various information sources
such as blogs~\cite{conrad2007,harb2008}, product
reviews~\cite{Hu04,Qadir2009,dave2003}, news
articles~\cite{kim2006,hu2006} etc. Various aspects in opinion
mining have been explored over the years~\cite{ku2006}. One important
dimension is to identify the opinion holders as well as opinion targets. \cite{Lu10} used
dependency parser to identify the opinion holders and targets in
Chinese news text. \cite{Choi05} use Conditional Random Fields to
identify the sources of opinions from the
sentences. \cite{kobayashi2005} propose a learning based anaphora
resolution technique to extract the opinion tuple
$<Subject,Attribute,Value>$. Opinion summarization has been another
important aspect~\cite{kim2013}.

A lot of research work has been done for 
opinion mining from product reviews where most of the text
is opinion-rich. Opinion mining from news articles, however, poses its
own challenges because in contrast with the product reviews, not all
parts of news articles present opinions~\cite{balahur2013} and thus
finding opinionated sentences itself remains a major obstacle. Our work mainly focus on classifying a sentence in a news article as
opinionated or factual. There have been works on sentiment
classification~\cite{wiebe2005} but the task of finding opinionated
sentences is different from finding sentiments, because sentiments
mainly convey the emotions and not the opinions. There has been
research on finding opinionated sentences from various
information sources. Some of these works utilize a dictionary-based~\cite{Fei12}
or regular pattern based~\cite{brun2012} approach to identify aspects
in the sentences. \cite{kim2006} utilize the presence of a single
strong valence wors as well as the total valence score of all words in a sentence to
identify opinion-bearing sentences. \cite{zhai11} work on finding `evaluative'
sentences in online discussions. They exploit the inter-relationship
of aspects, evaluation words and emotion words to reinforce each
other. 

Thus, while ours is not the first attempt at opinion extraction from
news articles, to the best of our knowledge, none of the previous works has
exploited the global structure of a news article to classify a
sentence as opinionated/factual. Though summarization algorithms~\cite{erkan2004,goyal2013}
utilize the similarity between sentences in an article to find the
important sentences, our formulation is different in that we conceptualize
two different kinds of nodes in a document, as opposed to the
summarization algorithms, which treat all the sentences equally. 

In the next section, we describe the propsoed two-stage algorithm in
detail.

\section{Our Approach}
\label{sec:method}
\begin{figure*}[thb]
\centering
  \includegraphics[width=12cm]{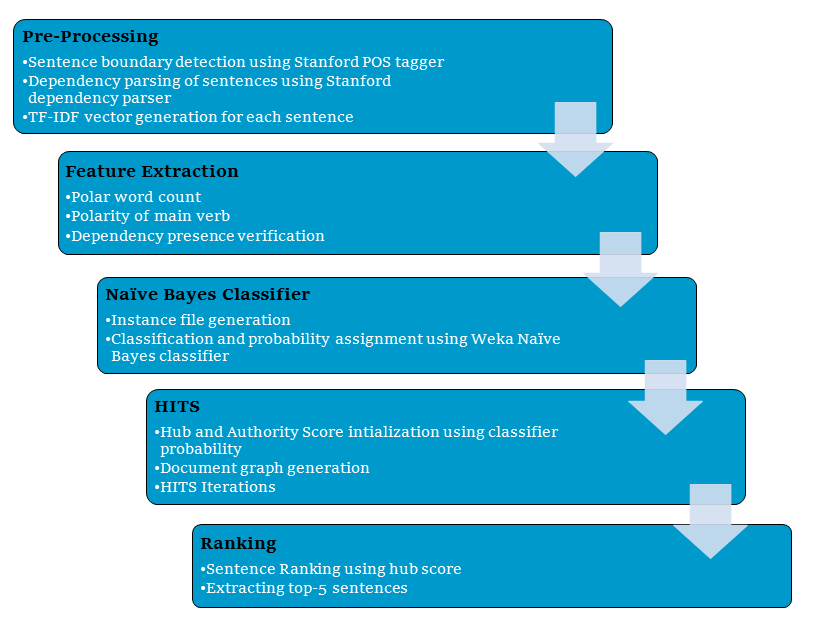}
  \caption{Flow Chart of Various Stages in Our Approach}
  \label{fig:flowchart}
\end{figure*}
Figure \ref{fig:flowchart} gives a flowchart of the proposed two-stage
method for extracting opinionated sentences from news
articles. First, each
news article is pre-processed to get the dependency parse as
well as the TF-IDF vector corresponding to each of the sentences
present in the article. Then, various features are extracted from
these sentences which are used as input to the Na\"{i}ve Bayes classifier, as will be
described in Section \ref{sec:nb}. The Na\"{i}ve Bayes
classifier, which corresponds to the first-stage of our method,
assigns a probability score to each sentence as being an
opinionated sentence. In the second
stage, the entire article is viewed as a complete and directed graph with
edges from every sentence to all other sentences, each edge having a weight suitably computed. Iterative HITS
algorithm is applied to the sentence graph, 
with opinionated sentences conceptualized as hubs and factual sentences conceptualized as
authorities. The two stages of our approach are detailed below.

\subsection{Na\"{i}ve Bayes Classifier}
\label{sec:nb}
The Na\"{i}ve Bayes classifier assigns the probability for each
sentence being opinionated. The classifier is trained on 70
News articles from politics domain, sentences of which were
marked by a group of annotators as being opinionated or factual. Each
sentence was marked by two annotators. The inter-annotator agreement
using Cohen's kappa coefficient was found to be $0.71$. 

The features utilized for the classifier are detailed in Table 1. These features were adapted from those
reported in~\cite{Qadir2009,Yu2003}.
A list of positive and negative polar words, further expanded using wordnet
synsets was taken from \cite{kim2005}. Stanford dependency
parser~\cite{de2006} was utilized to compute the dependencies for each
sentence within the news article. 

After the features are extracted from the sentences, we used the Weka
implementation of Na\"{i}ve Bayes to train the
classifier\footnote{\url{http://www.cs.waikato.ac.nz/ml/weka/}}. 
\begin{table}[h]
\begin{center}
\caption{Features List for the Na\"{i}ve Bayes Classifier}
\label{tab:features}
\begin{tabular}{|c|p{6cm}|}
\hline
1. & Count of positive polar words\\\hline
2. & Count of negative polar words\\\hline
3. & Polarity of the root verb of the sentence\\\hline
4. & Presence of aComp, xComp and advMod dependencies in the sentence\\\hline
\end{tabular}
\end{center}

\end{table}

\subsection{HITS}
\label{sec:hits}
The Na\"{i}ve Bayes classifier as discussed in Section \ref{sec:nb}
utilizes only the local features within a sentence. Thus, the
probability that a sentence is opinionated remains independent of its
context as well as the document structure. The main motivation behind
formulating this problem in HITS schema is to utilize the hidden link structures among sentences. 
HITS stands for `Hyperlink-Induced Topic
Search'; Originally, this algorithm
was developed to rank Web-pages, with a particular insight that some
of the webpages (\textbf{Hubs}) served as catalog of information, that could lead
users directly to the other pages, which actually contained the
information (\textbf{Authorities}).

The intuition behind applying HITS for the task of opinion extraction
came from the following assumption about underlying structure of an
article. A news article pertains to a specific theme and with that
theme in mind, the author presents certain opinions. These opinions are justified with
the facts present in the article itself. We conceptualize the
opinionated sentences as \textbf{Hubs} and the associated facts for an
opinionated sentence as \textbf{Authorities} for this \textbf{Hub}.

To describe the formulation of HITS parameters, let us give the
notations. 
Let us denote a document $D$ using a set of sentences
$\{S_1,S_2,\ldots,S_i,\ldots,S_n\}$, where $n$ corresponds to the number of
sentences in the document $D$. We construct the
sentence graph where nodes in the graph correspond to the sentences in the
document. Let $H_i$ and $A_i$ denote the hub and authority scores for
sentence $S_i$. In HITS, the edges always flow from a Hub to an
Authority. In the original HITS algorithm, each edge is given the
same weight. However, it has been reported that using weights
in HITS update improves the performance
significantly~\cite{li2002}. In our formulation, since each node has a
non-zero probablility of acting as a hub as well as an authority, we
have outgoing as well as incoming edges for every node. Therefore, the weights are
assigned, keeping in mind the proximity between sentences as well as the
probability (of being opinionated/factual) assigned by the classifier. The following
criteria were used for deciding the weight function.
\begin{itemize}
\item An edge in the HITS graph goes from a hub (source node) to an
  authority (target node). So, the edge weight from a source node to a
  target node should be higher if the source node has a high hub
  score.
\item A fact corresponding to an opinionated sentence should be
  discussing the same topic. So, the edge weight should be higher if
  the sentences are more similar.
\item It is more probable that the facts around an opinion appear
  closer to that opinionated sentence in the article. So, the edge weight from a source to target
  node decreases as the distance between the two sentences increases.
\end{itemize}

Let $W$ be the weight matrix such that $W_{ij}$ denotes the weight for
the edge from the sentence $S_i$ to the sentence $S_j$. Based on the
criteria outlined above, we formulate that the weight $W_{ij}$ should be
such that
\begin{eqnarray*}
W_{ij} &\propto& H_i\nonumber\\
W_{ij} &\propto& Sim_{ij}\nonumber\\
W_{ij} &\propto& \frac{1}{dist_{ij}}\nonumber
\end{eqnarray*}
where we use {\em cosine similarity} between the sentence vectors to
compute $Sim_{ij}$. $dist_{ij}$ is simply the number of sentences
separating the source and target node. Various combinations of these
factors were tried and will be discussed in section
\ref{sec:results}. While factors like sentence similarity and distance
are symmetric, having the weight function depend on the hub score
makes it asymmetric, consistent with the basic idea of HITS. Thus, an
edge from the sentence $S_i$ to $S_j$ is given a high weight if $S_i$
has a high probability score of being opinionated (i.e., acting as
hub) as obtained the classifier.

Now, for applying the HITS algorithm iteratively, the Hubs and
Authorities scores for each sentence are initialized using the
probability scores assigned by the classifier. That is, if
$P_i(Opinion)$ denotes the probability that $S_i$ is an opinionated
sentence as per the Na\"{i}ve Bayes Classifier, $H_i(0)$ is
initialized to $P_i(Opinion)$ and $A_i(0)$ is initialized to
$1-P_i(Opinion)$. The iterative HITS is then applied as follows:
\begin{eqnarray}
H_i(k) &=& \Sigma_j W_{ij} A_i(k-1)\\
A_i(k) &=& \Sigma_j W_{ji} H_i(k-1)
\end{eqnarray}

where $H_i(k)$ denote the hub score for the $i^{th}$ sentence during
the $k^{th}$ iteration of HITS. The iteration is stopped once the mean
squared error between the Hub and Authority values at two different
iterations is less than a threshold $\epsilon$. After the HITS iteration is over,
five sentences having the highest Hub scores are returned by the system. 

\section{Experimental Framework and Results}
\label{sec:results}

The experiment was conducted with 90 news articles in politics domain
from Yahoo! website. The sentences in the articles were marked as
opinionated or factual by a group of annotators. In the training set,
1393 out of 3142 sentences were found to be opinianated. In the test set, 347
out of 830 sentences were marked as opinionated. Out of these $90$ articles, $70$
articles were used for training the Na\"{i}ve Bayes classifier as
well as for tuning various parameters. The rest $20$ articles were used
for testing. The evaluation was done in an Information Retrieval
setting. That is, the system returns the sentences in a decreasing
order of their score (or probability in the case of Na\"{i}ve Bayes) as
being opinionated. We then utilize the human judgements (provided by
the annotators) to compute
precision at various points. Let $op(.)$ be a binary function for a
given rank such that $op(r)=1$ if the sentence returned as rank
$r$ is opinionated as per the human judgements. 

A $P@k$ precision is calculated as follows:
\begin{equation}
P@k=\frac{\sum_{r=1}^{k}op(r)}{k}
\end{equation}

While the precision at various points indicates how reliable the
results returned by the system are, it does not take into account 
the fact that some of the documents are opinion-rich and some are
not. For the opinion-rich documents, a high $P@k$ value might
be similar to picking sentences randomly, whereas for the documents
with a very few opinions, even a lower $P@k$ value might be useful. We,
therefore, devise another evaluation metric $M@k$ that indicates the ratio
of opinionated sentences at any point, normalized with respect to the
ratio of opinionated sentences in the article.

\noindent Correspondingly, an $M@k$ value is calculated as
\begin{equation}
M@k=\frac{P@k}{Ratio_{op}}
\end{equation}
where $Ratio_{op}$ denotes the fraction of opinionated sentences in the
whole article. Thus
\begin{equation}
Ratio_{op}=\frac{\text{Number of opinionated sentences}}{\text{Number of sentences}}
\end{equation}

The parameters that we needed to fix for the HITS algorithm were the
weight function $W_{ij}$ and the threshold $\epsilon$ at which we stop
the iteration. We varied $\epsilon$ from $0.0001$ to $0.1$
multiplying it by $10$ in each step. The results were not sensitive
to the value of $\epsilon$ and we used $\epsilon=0.01$. For fixing the weight
function, we tried out various combinations using the criteria
outlined in Section \ref{sec:hits}. Various weight functions and the
corresponding $P@5$ and $M@5$ scores are shown in Table
2. Firstly, we varied $k$ in ${Sim_{ij}}^k$ and
found that the square of the similarity function gives better
results. Then, keeping it constant, we varied $l$ in ${H_{i}}^l$ and
found the best results for $l=3$. Then, keeping both of these
constants, we varied $\alpha$ in $(\alpha+\frac{1}{d})$. We found the
best results for $\alpha=1.0$. With this $\alpha$, we tried to vary
$l$ again but it only reduced the final score. Therefore, we fixed the
weight function to be
\begin{equation}
\label{eq:weight}
W_{ij} = {H_i }^3(0){Sim_{ij}}^2(1+\frac{1}{dist_{ij}})
\end{equation}
Note that $H_i(0)$ in Equation \ref{eq:weight} corresponds to the
probablity assigned by the classifier that the sentence $S_i$ is
opinionated.

\begin{table}[!thb]
\begin{center}
\caption{Average $P@5$ and $M@5$ scores: Performance comparison
  between various functions for $W_{ij}$}
\begin{small}
\begin{tabular}{|l|c|c|}
\hline
Function & $P@5$ &  $M@5$ \\\hline
\textbf{$Sim_{ij}$} & 0.48 & 0.94\\\hline
\textbf{$Sim_{ij}^{2}$} & 0.57 & 1.16\\\hline
\textbf{$Sim_{ij}^{3}$} & 0.53 & 1.11\\\hline
\textbf{$Sim_{ij}^{2} H_{i}$} & 0.6 & 1.22\\\hline
\textbf{$Sim_{ij}^{2} {H_{i}}^2$} & 0.61 & 1.27\\\hline
\textbf{$Sim_{ij}^{2} {H_{i}}^{3}$} & 0.61 & 1.27\\\hline
\textbf{$Sim_{ij}^{2} {H_{i}}^{4}$} & 0.58 & 1.21\\\hline
\textbf{$Sim_{ij}^{2} {H_{i}}^{3} \frac{1}{d}$} & 0.56 & 1.20\\\hline
\textbf{$Sim_{ij}^{2} {H_{i}}^{3} (0.2 + \frac{1}{d})$} & 0.60 & 1.25\\\hline
\textbf{$Sim_{ij}^{2} {H_{i}}^{3} (0.4 + \frac{1}{d})$} & 0.61 & 1.27\\\hline
\textbf{$Sim_{ij}^{2} {H_{i}}^{3} (0.6 + \frac{1}{d})$} & 0.62 & 1.31\\\hline
\textbf{$Sim_{ij}^{2} {H_{i}}^{3} (0.8 + \frac{1}{d})$} & 0.62 & 1.31\\\hline
\textbf{$Sim_{ij}^{2} {H_{i}}^{3} (1 + \frac{1}{d})$} & \textbf{0.63} & \textbf{1.33}\\\hline
\textbf{$Sim_{ij}^{2} {H_{i}}^{3} (1.2 + \frac{1}{d})$} & 0.61 & 1.28\\\hline
\textbf{$Sim_{ij}^{2} {H_{i}}^{2} (1 + \frac{1}{d})$}  & 0.6 & 1.23\\\hline
\end{tabular}
\end{small}
\end{center}
\label{tab:sensitivity}
\end{table}


We use the classifier results as the baseline for the
comparisons. The second-stage HITS algorithm is then applied and we
compare the performance with respect to the classifier. Table 3 shows the comparison results for various
precision scores for the classifier and the HITS algorithm. In
practical situation, an editor requires quick identification of 3-5
opinionated sentences from the article, which she can then use to
formulate questions. We thus report $P@k$ and $M@k$ values for $k=3$
and $k=5$.

\begin{table}[!thb]
\label{tab:results}
\begin{center}
\caption{Average $P@5$, $M@5$, $P@3$ and $M@3$ scores: Performance comparison between
  the NB classifier and HITS}
\begin{tabular}{|l|c|c|c|c|}
\hline
System & P@5 &  M@5 & P@3 & M@3\\\hline
\textbf{NB Classifier} & 0.52 & 1.13 & 0.53 & 1.17\\\hline
\textbf{HITS} & 0.63 & 1.33 & 0.72 & 1.53\\\hline
\textbf{Imp. (\%)} & \textbf{+21.2} & \textbf{+17.7} &
\textbf{+35.8} & \textbf{+30.8}\\\hline
\end{tabular}
\end{center}
\end{table}

From the results shown in Table 3, it is clear that
applying the second-stage HITS over the Na\"{i}ve Bayes Classifier
improves the performance by a large degree, both in term of $P@k$ and
$M@k$. For instance, the first-stage NB Classifier gives a $P@5$ of
$0.52$ and $P@3$ of $0.53$. Using the classifier outputs during the second-stage HITS
algorithm improves the preformance by $21.2\%$ to $0.63$ in the case
of $P@5$. For $P@3$, the improvements were much more significant and a
$35.8\%$ improvement was obtained over the NB classifier. $M@5$ and
$M@3$ scores
also improve by $17.7\%$ and $30.8\%$ respectively. 

Strikingly, while the classifier gave nearly the same scores for $P@k$
and $M@k$ for $k=3$ and $k=5$, HITS gave much better results for $k=3$
than $k=5$. Specially, the $P@3$ and $M@3$ scores
obtained by HITS were very encouraging, indicating that the
proposed approach helps in pushing the opinionated sentences to the
top. This clearly shows the advantage of using
the global structure of the document in contrast with the features
extracted from the sentence itself, ignoring the context. 

\begin{figure*}[thb]
\label{fig:comp}
\centering
\begin{subfigure}{.5\textwidth}
  \centering
  \includegraphics[width=7.5cm]{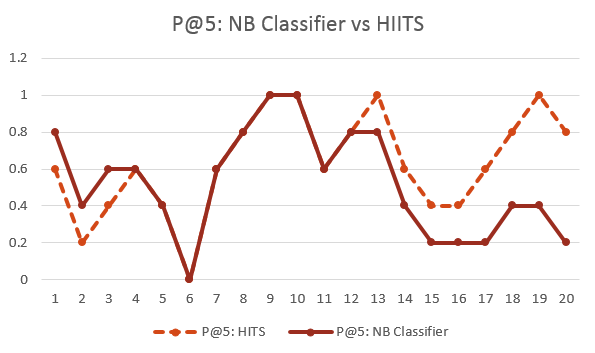}
  \caption{Comparison of P@5 values}
  \label{fig:sub1}
\end{subfigure}%
\begin{subfigure}{.5\textwidth}
  \centering
  \includegraphics[width=7.5cm]{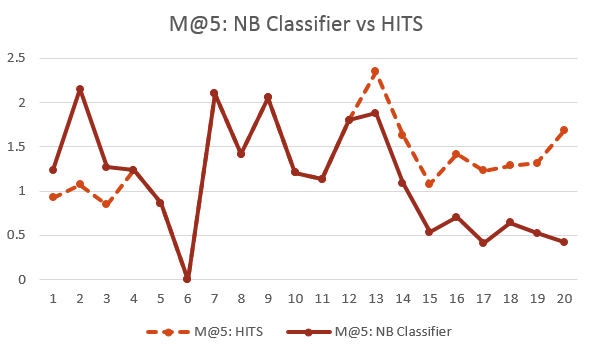}
  \caption{Comparison of M@5 values}
  \label{fig:sub2}
\end{subfigure}
\caption{Comparison Results for 20 Test articles between the Classifier
and HITS: P@5 and M@5}
\end{figure*}

\begin{figure*}[thb]
\label{fig:comp}
\centering
\begin{subfigure}{.5\textwidth}
  \centering
  \includegraphics[width=7.5cm]{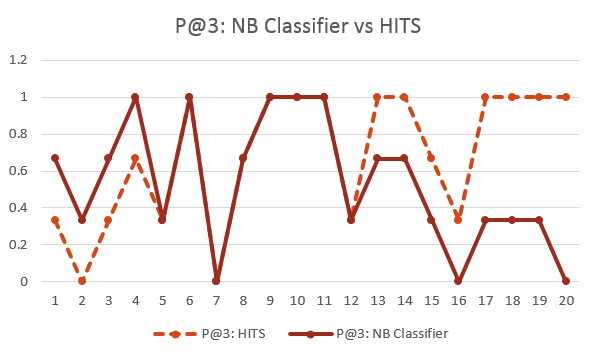}
  \caption{Comparison of P@3 values}
  \label{fig:sub1}
\end{subfigure}%
\begin{subfigure}{.5\textwidth}
  \centering
  \includegraphics[width=7.5cm]{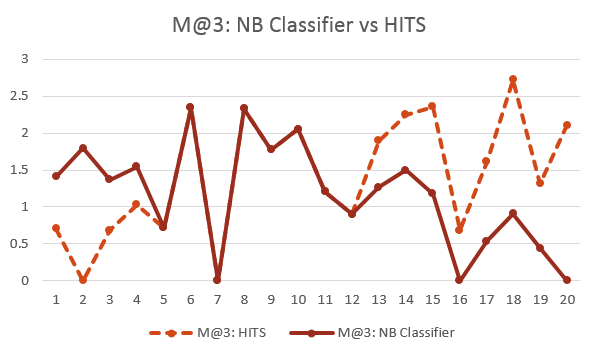}
  \caption{Comparison of M@3 values}
  \label{fig:sub2}
\end{subfigure}
\caption{Comparison Results for 20 Test articles between the Classifier
and HITS: P@3 and M@3}
\end{figure*}

Figures 2 and 3 show the $P@5$, $M@5$, $P@3$ and $M@3$ scores for individual
documents as numbered from $1$ to $20$ on the X-axis. The articles
are sorted as per the ratio of $P@5$ (and $M@5$)
obtained using the HITS and NB classifier. Y-axis shows the
corresponding scores. Two different lines are used to represent the results as
returned by the classifier and the HITS algorithm. A dashed line denotes
the scores obtained by HITS while a continuous line denotes the scores
obtained by the NB classifier. A detailed analysis of these figures can
help us draw the following conclusions:
\begin{itemize}
\item For $40\%$ of the articles (numbered $13$
to $20$) HITS improves over the baseline NB classifier. For $40\%$ of
the articles (numbered $5$ to $12$) the results provided by
HITS were the same as that of the baseline. For $20\%$ of the articles
(numbered $1$ to $4$) HITS gives a performance lower than that of the
baseline. Thus, for $80\%$ of the documents, the second-stage performs
at least as good as the first stage. This indicates that the
second-stage HITS is quite robust.
\item $M@5$ results are much more robust for the HITS, with $75\%$ of
  the documents having an $M@5$ score $>1$. An $M@k$ score $>1$
  indicates that the ratio of opinionated sentences in top $k$
  sentences, picked up by the algorithm, is higher than the overall
  ratio in the article.
\item For $45\%$ of the articles, (numbered $6$, $9-11$ and $15-20$),
  HITS was able to achieve a $P@3=1.0$. Thus, for these 9 articles,
  the top $3$ sentences picked up by the algorithm were all marked as
  opinionated.
\end{itemize}

The graphs also indicate a high correlation between the results
obtained by the NB classifier and HITS. We used Pearson's correlation
to find the correlation strength. For the $P@5$ values, the correlation
was found to be $0.6021$ and for the $M@5$ values, the correlation was
obtained as $0.5954$.

In the next section, we will first attempt to further analyze the
basic assumption behind using HITS, by looking at some actual Hub-Authority
structures, captured by the algorithm. We will also take some
cases of failure and perform error analysis.

\section{Discussion}
\label{sec:discuss}

First point that we wanted to verify was, whether
HITS is really capturing the underlying structure of the document. That is,
are the sentences identified as authorities for a given hub really
correspond to the facts supporting the particular opinion, expressed
by the hub sentence.

\begin{figure*}[!thb]
\label{fig:hub_auth}
\centering
\begin{subfigure}{.5\textwidth}
  \centering
  \includegraphics[width=8cm]{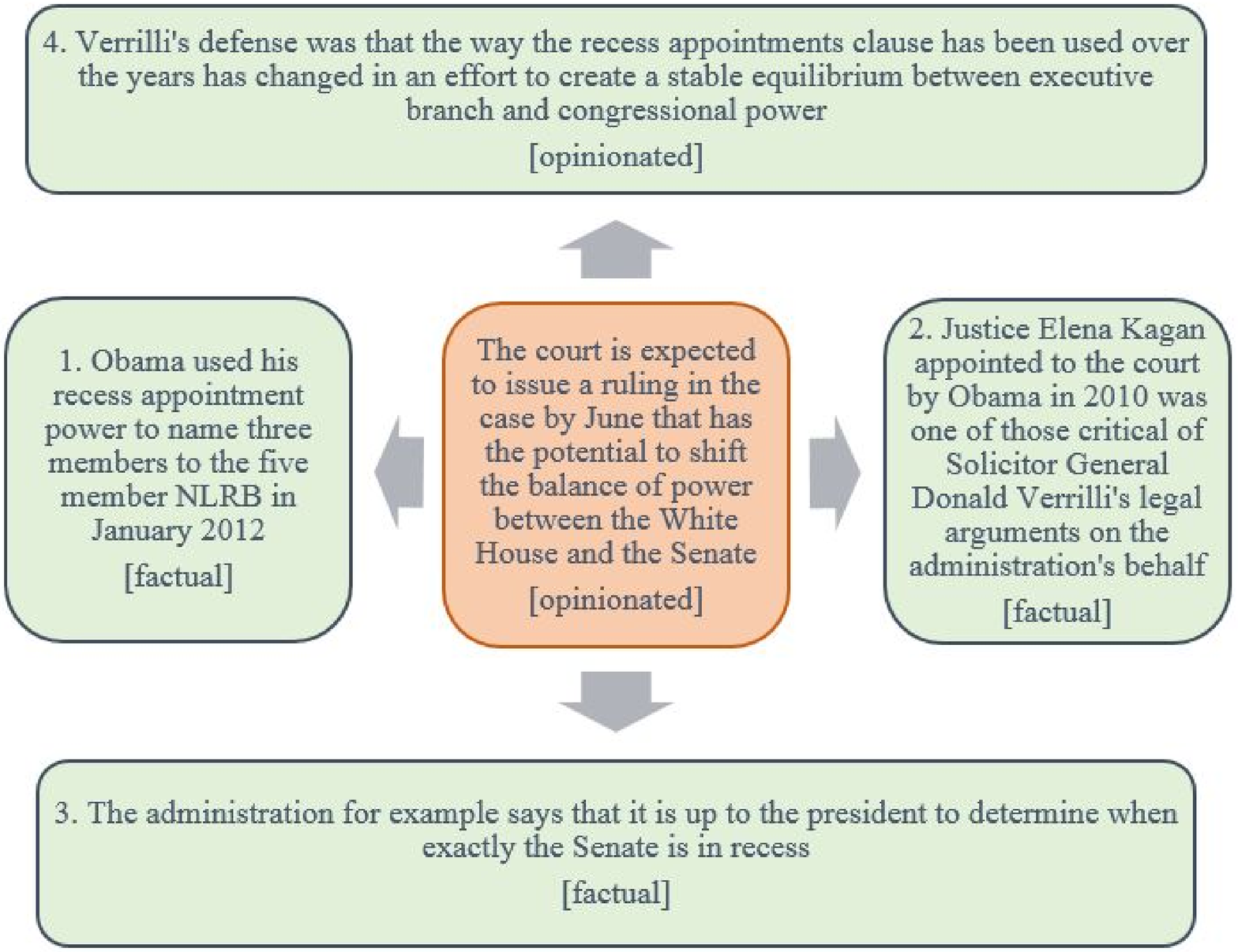}
  \caption{Hub-Authority Structure: Example 1}
  \label{fig:sub1}
\end{subfigure}%
\begin{subfigure}{.5\textwidth}
  \centering
  \includegraphics[width=8cm]{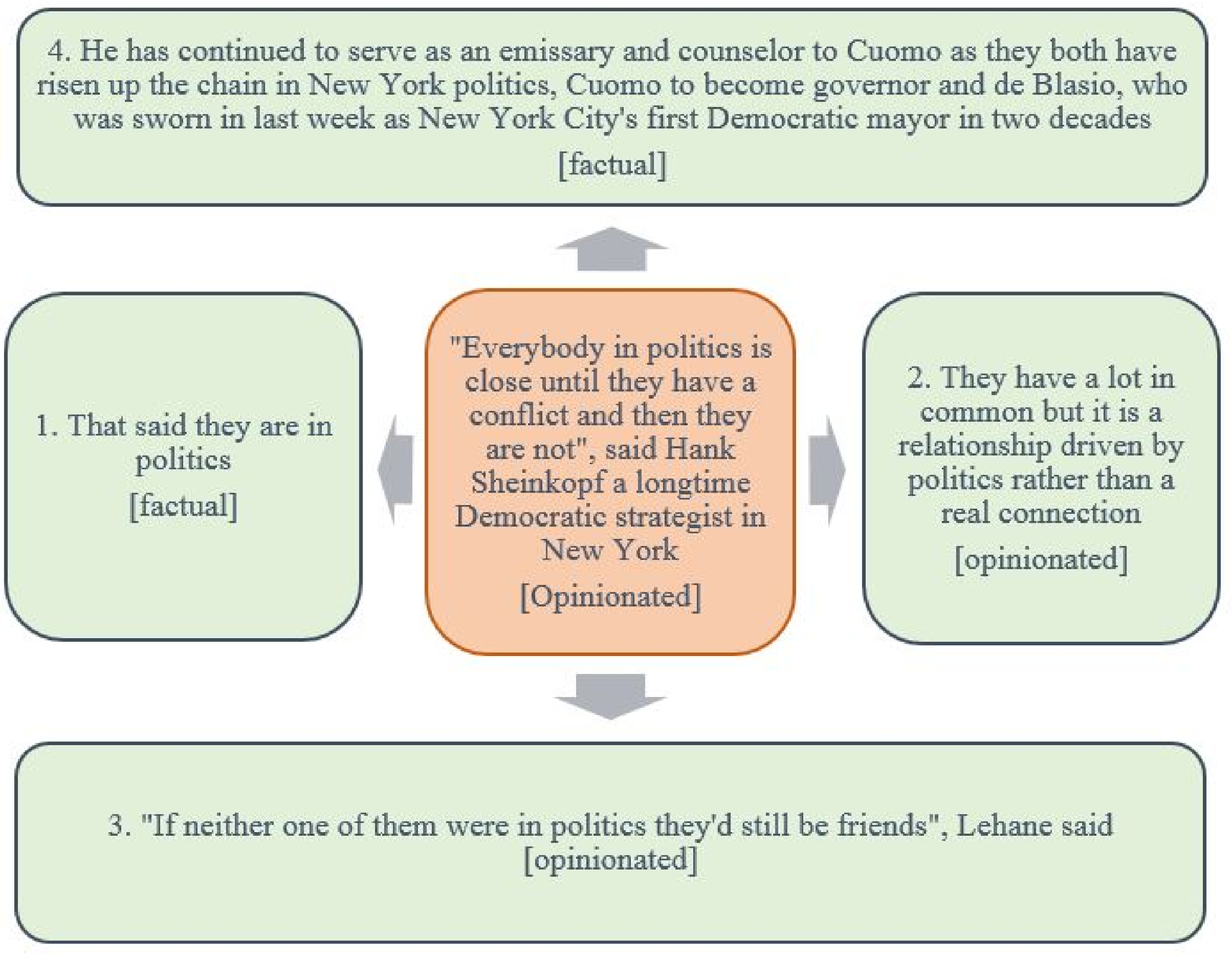}
  \caption{Hub-Authority Structure: Example 2}
  \label{fig:sub2}
\end{subfigure}
\caption{Example from two different test articles capturing the Hub-Authority Structure}

\end{figure*}

Figure 4 gives two examples of the Hub-Authority structure,
as captured by the HITS algorithm, for two different articles. For each
of these examples, we show the sentence identified as Hub in the
center along with the top four sentences, identified as Authorities for
that hub. We also give the annotations as to whether the sentences
were marked as `opinionated' or `factual' by the annotators. 

In both of these examples, the hubs were actually marked as
`opinionated' by the annotators. Additionally, we find that all the
four sentences, identified as authorities to the hub, are very
relevant to the opinion expressed by the hub. In the first example,
top 3 authority sentences are marked as `factual' by the
annotator. Although the fourth sentence is marked as `opinionated', it
can be seen that this sentence presents a
supporting opinion for the hub sentence.

While studying the second example, we found that while the first
authority does not present an important fact, the fourth authority
surely does. Both of these were marked as `factual' by the annotators.
In this particular example, although the second and third authority
sentences were annotated as `opinionated', these can be seen as
supporting the opinion expressed by the hub sentence. This example
also gives us an interesting idea to improve
diversification in the final results. That is, once an opinionated sentence is identified
by the algorithm, the hub score of all its suthorities can be
reduced proportional to the edge weight. This will reduce the chances
of the supporting opinions being reurned by the system, at a later
stage as a main opinion.

We then attempted to test our tool on a recently published article,
``\textbf{What's Wrong with a  Meritocracy
Rug?}''\footnote{\url{http://news.yahoo.com/whats-wrong-meritocracy-rug-070000354.html}}. The
tool could pick up a very important opinion in the article, ``{\sl Most
people tend to think that the most qualified person is someone who
looks just like them, only younger.}'', which was ranked $2^{nd}$ by
the system. The supporting facts and opinions for this sentence, as
discovered by the algorithm were also quite relevant. For instance, the top
two authorities corresponding to this sentence hub were:
\begin{enumerate}
\item {\sl And that appreciation, we learned painfully, can easily be
  tinged with all kinds of gendered elements without the person who is
  making the decisions even realizing it.}
\item {\sl And many of the traits we value, and how we value them, also end
  up being laden with gender overtones.}
\end{enumerate}

\subsection{Error Analysis}
We then tried to analyze certain cases of failures. Firstly, we wanted
to understand why HITS was not performing as good as the classifier
for 3 articles (Figures 2 and 3). The analysis revealed that the supporting
sentences for the opinionated sentences, extracted by the classifier,
were not very similar on the textual level. Thus a low cosine
similarity score resulted in having lower edge weights, thereby
getting a lower hub score after applying HITS. For one of the
articles, the sentence picked up by HITS was wrongly annotated as
a factual sentence.

Then, we looked at one case of failure due to the error introduced by
the classifier prior probablities. For instance, the
sentence, ``{\sl The civil war between establishment and tea party
  Republicans \textbf{intensified} this week when House Speaker John Boehner
  slammed outside \textbf{conservative} groups for \textbf{ridiculous} pushback against
  the bipartisan budget agreement which cleared his chamber
  Thursday.}'' was classified as an opinionanted sentence, whereas
this is a factual sentence. Looking closely, we found that the
sentence contains three polar words (marked in bold), as well as an
$advMod$ dependency between the pair (slammed,when). Thus the sentence
got a high initial prior by the classifier. As a result, the outgoing
edges from this node got a higher ${H_i}^3$ factor. Some of the authorities
identified for this sentence were:
\begin{itemize}
\item {\sl For Democrats, the tea party is the gift that keeps on giving.}
\item {\sl Tea party sympathetic organizations, Boehner later said, ``are
  pushing our members in places where they don't want to be".}
\end{itemize}
which had words, similar to the original sentence, thus having a
higher $Sim_{ij}$ factor as well. We found that these sentences were
also very close within the article. Thus, a high hub prior along with
a high outgoing weight gave rise to this sentence having a high hub
score after the HITS iterations.

\begin{figure*}[!thb]
 \centering
  \includegraphics[width=13cm]{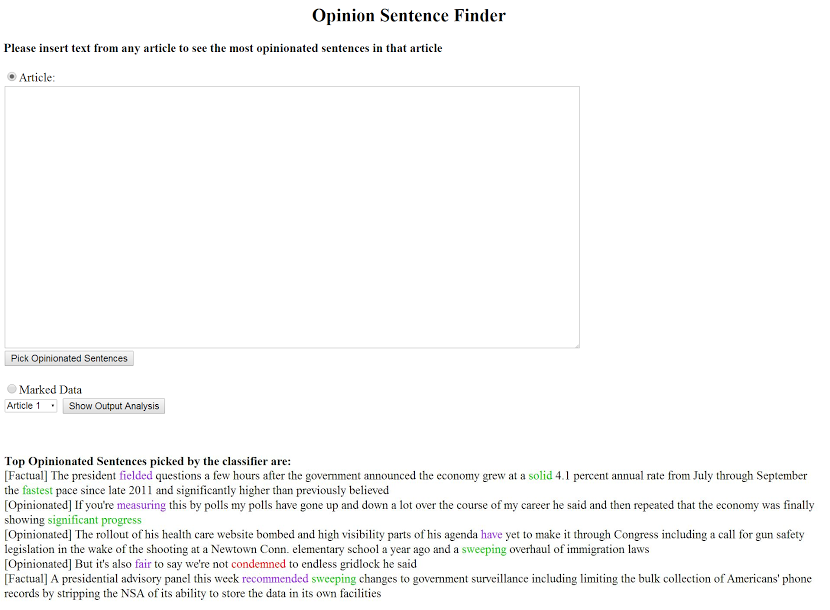}
  \caption{Screenshot from the Web Interface}
  \label{fig:screen1}
\end{figure*}

\begin{figure*}[!thb]
 \centering
  \includegraphics[width=13cm]{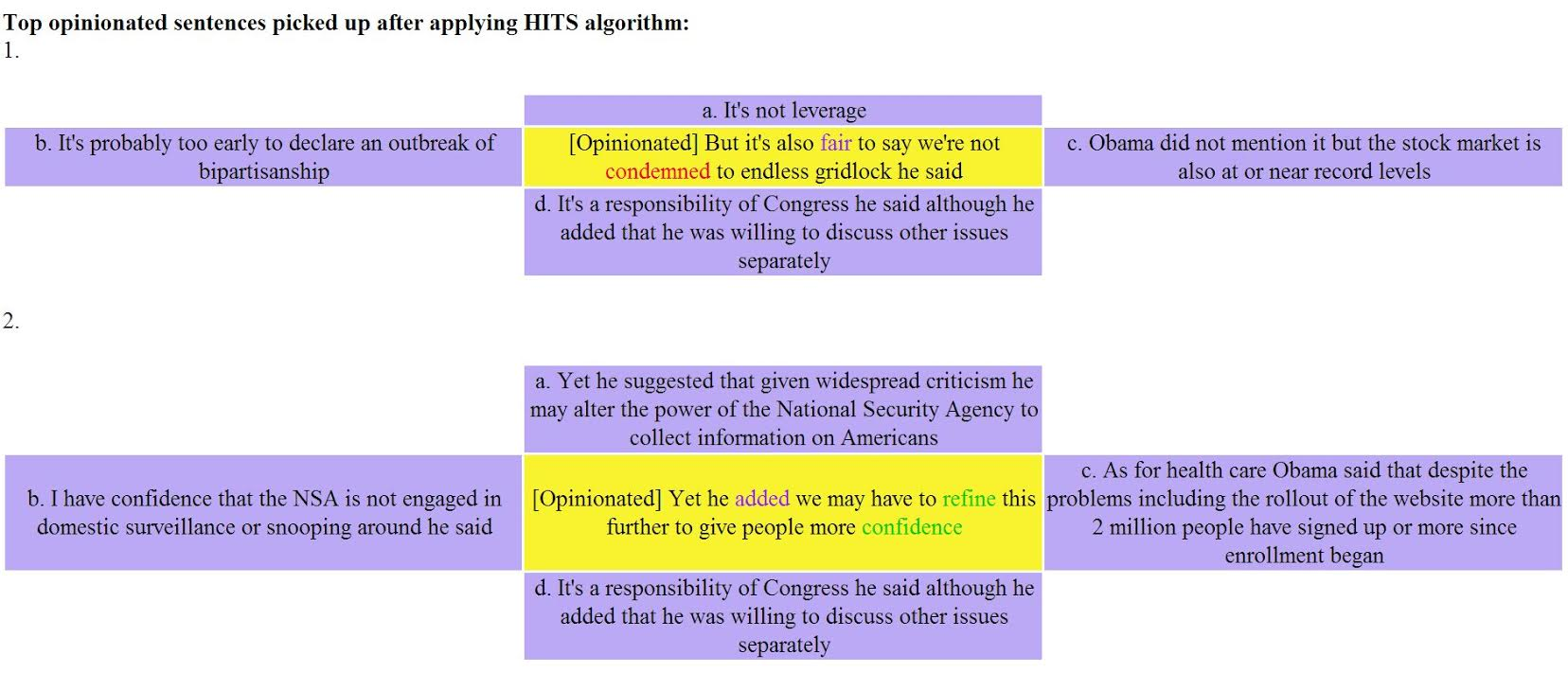}
  \caption{Hub-Authority Structure as output on the Web Interface}
  \label{fig:screen2}
\end{figure*}

\subsection{Online Interface}
To facilitate easy usage and understanding of the system by others, a web interface has
been built for the system\footnote{available at \url{http://cse.iitkgp.ac.in/resgrp/cnerg/temp2/final.php}}. The webpage caters for users to either
input a new article in form of text to get top opinionated sentences or view the output analysis of the system over manually
marked test data consisting of 20 articles.

The words in green color are positive polar words, red indicates negative polar words.
Words marked in violet are the root verbs of the sentences. The colored graph shows top
ranked opinionated sentences in yellow box along with top supporting factual sentences
for that particluar opinionated sentence in purple boxes. Snapshots from the online interface
are provided in Figures 5 and 6.

\section{Conclusions and Future Work}
\label{sec:future}
In this paper, we presented a novel two-stage framework for extracting
the opinionated sentences in the news articles. The problem of
identifying top opinionated sentences from news articles is very
challenging, especially because the opinions are not as explicit in a
news article as in a discussion forum. It was also evident from the
inter-annotator agreement and the kappa coefficient was found to be
$0.71$.

The experiments conducted
over 90 News articles (70 for training and 20 for testing) clearly indicate that the proposed two-stage
method almost always improves the performance of the baseline
classifier-based approach. Specifically, the improvements are much
higher for $P@3$ and $M@3$ scores ($35.8\%$ and $30.8\%$ over the NB
classifier). An $M@3$ score of $1.5$ and $P@3$ score of $0.72$
indicates that the proposed method was able to push the opinionated
sentences to the top. On an average, $2$ out of top $3$ sentences
returned by the system were actually opinionated. This is very much
desired in a practical scenario, where an editor requires quick
identification of 3-5 opinionated sentences, which she can then use to
formulate questions.

The examples discussed in Section \ref{sec:discuss} bring out another
important aspect of the proposed algorithm. In addition to the main
objective of extracting the opinionated sentences within the article,
the proposed method actually discovers the underlying structure of the
article and would certainly be useful to present various opinions,
grouped with supporting facts as well as supporting opinions in the
article.

While the initial results are encouraging, there is scope for
improvement. We saw that the results obtained via HITS were
highly correlated with the Na\"{i}ve Bayes classifier results, which
were used in assigning a weight to the document graph. One
direction for the future work would be to experiment with other
features to improve the precision of the classifier. Additionally, in
the current evaluation, we are not evaluating the degree of diversity
of the opinions returned by the system. The Hub-Authority structure of
the second example gives us an interesting idea to improve
diversification and we would like to implement that in future.

In the future, we would also like to apply this work to track an event over
time, based on the opinionated sentences present in the articles. When
an event occurs, articles start out with more factual sentences. Over
time, opinions start surfacing on the event, and as the event matures,
opinions predominate the facts in the articles. For example, a set of
articles on a plane crash would start out as factual, and would offer
expert opinions over time. This work can be used to plot the maturity
of the media coverage by keeping track of facts v/s opinions on any
event, and this can be used by organizations to provide a timeline for
the event. We would also like to experiment with this model on a
different media like microblogs.


\end{document}